\documentclass[conference]{IEEEtran}
\usepackage[T1]{fontenc}
\usepackage[utf8]{inputenc}
\usepackage{amsmath,amssymb}
\usepackage{booktabs}
\usepackage{array,tabularx,multirow}
\usepackage{graphicx}
\usepackage{xcolor}
\usepackage{tikz}
\usetikzlibrary{arrows.meta,positioning,fit,calc}
\usepackage{microtype}
\usepackage[hyphens]{url}
\usepackage[hidelinks]{hyperref}

\newcolumntype{Y}{>{\raggedright\arraybackslash}X}
\newcommand{\code}[1]{\texttt{#1}}
\newcommand{\histnote}{\textsuperscript{\dag}}

\title{Compute Regeneration through Re-Productization:\\Reorganizing Retired Edge AI Silicon into a\\Delivered Multilingual Service Robot}

\IEEEoverridecommandlockouts
\author{\IEEEauthorblockN{Hongyu Luo}
\IEEEauthorblockA{Beijing Data Chain\\
Technology Co., Ltd.\\
Beijing, China\\
ORCID: 0009-0002-0152-2970\\
Email: luohongyu@gravity.link}
\and
\IEEEauthorblockN{Sun Tie}
\IEEEauthorblockA{University of Science\\
and Technology Beijing\\
Beijing, China\\
\mbox{}\\
Email: ggw@ustb.edu.cn}
\and
\IEEEauthorblockN{John Wang}
\IEEEauthorblockA{Beijing Data Chain\\
Technology Co., Ltd.\\
Beijing, China\\
\mbox{}\\
Email: john.wang@gravity.link}
\thanks{Evidence-status draft revised on 2026-11-07. The October 2026 performance values are transcribed from the first author's Extended Abstract v1.0; their original scripts and JSON logs have not been recovered and are not claimed as reproduced. A separate, independently auditable November campaign was run on the same Xavier after restoring an isolated CUDA/PyTorch stack; it is reported in parallel with, and never pooled with, the historical values. The trace-driven analysis of the finite-pulse execution boundary is derived from measured host traces and the audited firmware contract; it is not a physical measurement. The motor UART is online and was exercised with inert no-motion frames, but fault injection during motion, the six-run product demonstration, and logic-analyser capture of the MCU pulse pins remain unresolved; non-physical, software-in-the-loop product-loop and six-mode full-runtime control-state replays are reported separately and are not substitutes for them. Drafting, source inspection, and test scaffolding were assisted by AI tools; the authors are responsible for the experimental record and technical decisions.}}

\begin{document}
\maketitle

\begin{abstract}
Useful edge hardware can lose product support while retaining substantial computation. We study this gap as a systems-design problem rather than a refurbishment problem. \emph{Compute regeneration} is defined here as re-productization: retaining the compute module while redesigning the division of labor, perception interfaces, and product function around it. The case artifact is GEdgeBot, a wheeled multilingual service robot built around a retired 32~GB Jetson AGX Xavier. The product value is \emph{joint}, not a property of the module: a source-grounded integration inventory shows that the box alone supplies inference, whereas the combined system exposes five on-device AI services, a seven-language speech chain, three sensing domains, and one deterministic seven-byte control boundary. Its architecture assigns non-deterministic AI and orchestration to the Xavier, finite motor-pulse generation to an STM32F407, and sensing to a reverse-engineered microphone array, an OpenNI depth camera, and a separately integrated automotive FPD-Link camera path. A source-grounded analysis traces the seven-byte control boundary, the six-mode overall control state machine with its prewarm/readiness barriers, sound-orientation mapping, counter-turning head coordination, two-level visual following, model prewarming, and peripheral registration mechanisms. Historical single-prototype measurements report 1,782~MB GPU memory with YOLO, Whisper-small, and NLLB-200 resident; a rise in direct-control p99 timing deviation from 0.24 to 49.93~ms under concurrent AI load; and six of six completed trigger-to-follow demonstrations with a reported mean latency of 4.35~s, but with conversational voice unavailable in those demonstrations. New tests verify 17 protocol/logic contracts, and a separate November campaign on the same Xavier restores an isolated inference stack, re-measures the AI workloads and their parameter sensitivity, and repeats idle/AI-loaded control-frame runs five times per power mode. Driving two execution organisations with the same 6,000 measured 50~ms frame intervals shows what the finite-pulse boundary does and does not isolate: it removes host jitter from the pulse train but not from the command start phase, and its stopping bound is enforced by the host rather than by the MCU, whose 16-bit pulse field is unclamped. A six-check static audit exposes a host--MCU queue mismatch in the counter-turn sequence. A 14-scenario full-runtime replay executes the audited \code{main()} dispatcher across all six modes and ten complete auto-wakeup cycles with zero physical device opens and 54 contract-checked frames. These results support an existence claim and a reproducible architectural hypothesis, not fleet-level reliability, universal offline operation, or measured environmental benefit. The historical $208\times$ direct-control ratio remains unreplicated; we explicitly identify missing raw logs, source drift, protocol inconsistencies, and the MCU-side waveform capture and on-site fault injection still required.
\end{abstract}


\section{Introduction}
The physical utility of a compute module and the lifetime of its software support are not the same quantity. The distinction is especially visible in edge AI, where a module can retain useful acceleration while newer operating-system, compiler, and inference stacks move beyond its last supported board-support package. The global waste context makes this distinction important: the Global E-waste Monitor reports 62~million tonnes of e-waste generated in 2022 and 22.3\% formally collected and recycled~\cite{global_ewaste_monitor_2024}. Those figures motivate examination of reuse, but they do not by themselves establish a carbon or economic benefit for any particular device.

This paper examines a retired Jetson AGX Xavier as a concrete instance. Xavier provides substantial edge-AI computation, while NVIDIA's JetPack support places it on a different and older software track than newer Jetson modules~\cite{nvidia_jetson_agx_xavier,nvidia_jetpack_5,nvidia_jetpack_6}. The research question is therefore not whether an old board can execute another benchmark. It is whether the board can become the compute core of a different, delivered product by reorganizing the architecture around its residual capabilities and constraints.

We call this process \emph{compute regeneration through re-productization}. Refurbishment restores an existing product role. Repurposing runs a new workload in an inherited system. Re-productization instead treats the compute module as one asset in a new system design: fast-changing, non-deterministic workloads remain on the old accelerator; timing-critical execution is moved behind a narrow interface to a deterministic controller; peripherals are selected and integrated for the new product rather than for the module's original role.

The case artifact is GEdgeBot, a wheeled service robot integrating sound-source localization, person following, speech recognition, and multilingual translation around one retired Xavier. Its value as a research object is not a claim that every component is novel. It is that the complete system exposes the integration work that a component-level reuse study omits: control-domain separation, microphone protocol recovery, model-lifecycle management, and camera/BSP integration.

\textbf{Why the combination matters.} A retired Jetson box on its own is a compute asset with an expiring software-support trajectory: it can execute models, but has no sensing, no actuator, and no product role. The captured source README names the delivered artifact plainly---``wheel robot with GEdgeUltral AI box''---and shows what the combination adds: five active on-device AI services, a seven-language ASR/MT/TTS configuration with three consistent mapping tables, three sensing domains, two kernel driver modules for an automotive camera path, and a seven-byte control boundary to an STM32F407. Audited against the captured source (Section~\ref{sec:integration}), only one of six product functions is available from the box alone. The value is therefore joint rather than additive: it arises from the division of labour and the narrow interface between the old compute core and the newly designed product organisation. We make this gain auditable as an engineering inventory and deliberately do not assign it a monetary or carbon value (Section~\ref{sec:interpret}).

The paper makes four contributions:
\begin{enumerate}
\item It defines a four-rung value-capture model---recycle, refurbish, repurpose, and re-productize---and converts it into three architectural axes: evolution speed, determinism/safety, and peripheral domain.
\item It documents a source-grounded robot architecture, including the exact control-frame format, the six-mode overall control state machine, and orchestration mechanisms that are often reduced to a block labeled ``integration,'' and makes the resulting product-level capability gain auditable as a six-function integration inventory.
\item It reports the October 2026 single-artifact measurements with explicit evidence status, including degraded voice operation in the six demonstration runs and an unresolved inconsistency in the NLLB timing record.
\item It adds a November 2026 reproducibility audit: 17 passing source-contract tests, a new 200-interval no-motion timing run, an independently rerun AI measurement campaign on a restored isolated CUDA stack, and a trace-driven isolation analysis that separates command-start jitter from pulse-train jitter and identifies an unclamped MCU pulse field. A source-grounded audit and a 14-scenario full-runtime replay cover the complete six-mode control plane. A discrepancy register records the host--MCU counter-turn queue mismatch.
\end{enumerate}

The study deliberately stops short of claims requiring multiple devices, deployment statistics, a verified bill of materials, or lifecycle assessment. Its strongest claim is an existence proof accompanied by a falsifiable architectural explanation.

\section{Background and Related Work}
\subsection{Reuse is not one intervention}
Material recycling destroys the information-bearing organization of a chip in order to recover material. Refurbishment retains the product and restores its original function. Repurposing retains a system but changes its workload. Re-productization changes the system organization and product role while retaining the compute module. These categories can overlap in a particular bill of materials; the distinction used here is the \emph{object of design}, not a claim that every other component must be new.

DumpsterCluster evaluates discarded GPUs as a large-scale inference resource, asking whether old accelerators can still provide datacenter computation~\cite{dumpstercluster_2026}. Junkyard Computing similarly explores the use of discarded smartphones to minimize carbon~\cite{junkyard_computing}. GEdgeBot addresses a complementary scale and question: not aggregate compute from many discarded devices, but the architectural reorganization needed to deliver one new product from one old compute module.

\subsection{The software ceiling as an architectural constraint}
A frozen board-support package affects more than installation convenience. Compiler support, Python versions, accelerator libraries, model exporters, and upstream APIs co-evolve. A system built below that ceiling must pin working combinations, isolate components that require different runtimes, and sometimes carry local compatibility patches. The appropriate architectural response is therefore workload decomposition, not merely an attempt to install the newest framework.

For GEdgeBot, the relevant constraint is the Xavier/JetPack~5 software track rather than an assertion that the silicon has failed. The paper does not use unverified supply-chain or last-order dates as evidence. It uses the observable support boundary and the engineering work required to operate within it.

\subsection{Heterogeneous robot computation}
Service robots combine non-deterministic perception with timing-sensitive actuation. Running both on a general-purpose operating system creates interference paths through CPU scheduling, memory, preprocessing, and accelerator contention. A heterogeneous design can place AI inference on an accelerator while a microcontroller executes bounded actuation primitives. The safety boundary then depends on what crosses the interface: a persistent velocity command creates a different failure mode from a finite pulse command that terminates automatically.

GEdgeBot uses the latter. This choice does not make the system universally safe; it removes one runaway state while leaving protocol integrity, mechanical limits, and fault detection as separate problems.

\section{The Re-Productization Model}
\begin{figure}[t]
\centering
\begin{tikzpicture}[font=\small, box/.style={draw, align=center, inner sep=4pt}, flow/.style={-{Latex[length=2mm]}, thick}]
\node[box, fill=gray!12, minimum width=17mm, minimum height=12mm] (recycle) at (0,0) {Recycle\\material};
\node[box, fill=blue!6, minimum width=17mm, minimum height=18mm] (refurbish) at (1.85,0.35) {Refurbish\\same role};
\node[box, fill=green!7, minimum width=17mm, minimum height=24mm] (repurpose) at (3.7,0.7) {Repurpose\\new task};
\node[box, fill=orange!12, minimum width=20mm, minimum height=30mm] (reproduct) at (5.75,1.15) {Re-productize\\new system};
\draw[flow] (recycle) -- (refurbish);
\draw[flow] (refurbish) -- (repurpose);
\draw[flow] (repurpose) -- (reproduct);
\node[below=2mm of recycle, font=\footnotesize] {silicon destroyed};
\node[below=2mm of refurbish, font=\footnotesize, text width=20mm, align=center] {OEM limits retained};
\node[below=2mm of repurpose, font=\footnotesize, text width=22mm, align=center] {product constraints retained};
\node[below=2mm of reproduct, font=\footnotesize, text width=25mm, align=center] {only the compute core retained};
\end{tikzpicture}
\caption{The proposed value-capture ladder is a systems-design distinction, not a measured ranking of economic or environmental benefit. Re-productization preserves the compute module while redesigning the product organization around it; in this case an AI box becomes the compute core of a delivered service robot.}
\label{fig:ladder}
\end{figure}
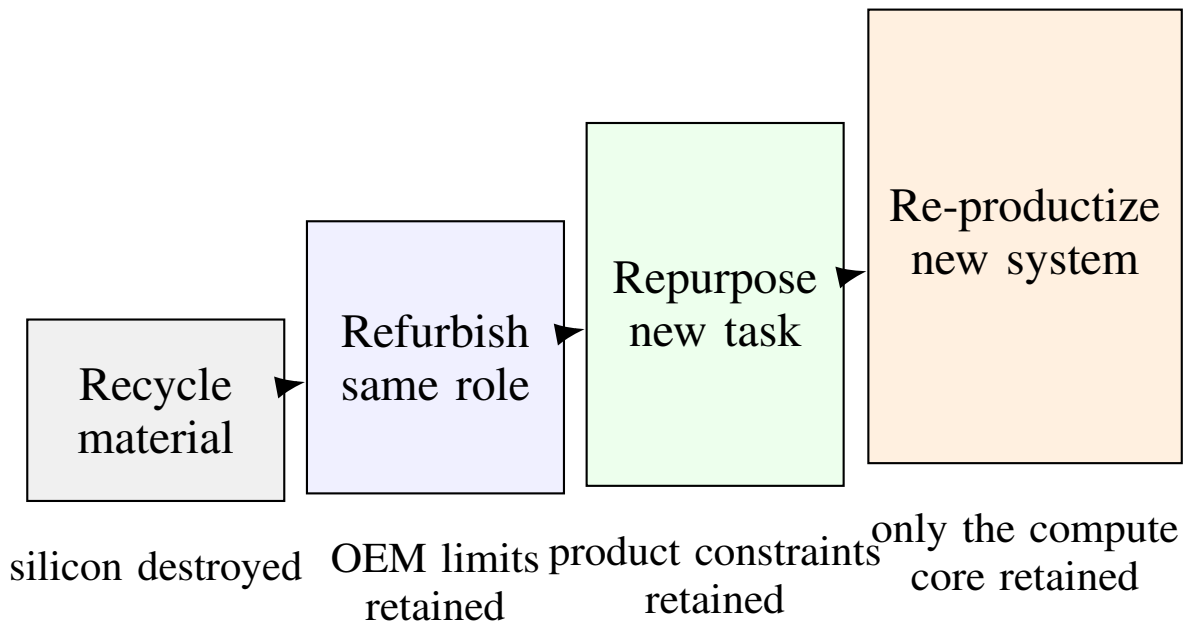

\subsection{What the combination adds}
\label{sec:integration}
Table~\ref{tab:integration} turns the value ladder into an auditable inventory. It is derived by parsing read-only source snapshots captured from the robot working tree (Git HEAD \code{c25b444}) together with the November device JSONs; it is a \emph{source-grounded inventory}, not a runtime measurement and not an economic assessment.

\begin{table}[t]
\centering
\caption{Product functions available from the retired compute module alone versus the module embedded in the delivered robot. Source-grounded integration inventory, derived from the captured source tree and the November device JSONs.}
\label{tab:integration}
\footnotesize
\begin{tabularx}{\columnwidth}{@{}p{34mm}p{16mm}Y@{}}
\toprule
Product function & Box alone & Box + robot: what the integration adds\\
\midrule
On-device inference (YOLO / Whisper / NLLB / ChatTTS) & yes\histnote & retained as the compute core; isolated CUDA stack on the frozen BSP\\
Seven-language ASR$\to$MT$\to$TTS & models only & five on-device services; microphone array, prewarm/readiness, orchestration\\
Sound orientation and counter-turn & no & recovered DOA protocol and seven-byte UART frames\\
Two-level person following & no & OpenNI camera and head/body pulse state machine\\
FPD-Link camera registration & no & two kernel driver modules and a V4L2 graph\\
Finite-pulse bounded actuation & no & STM32F407 and the seven-byte frame\\
\bottomrule
\end{tabularx}
\vspace{2pt}
\begin{flushleft}
\footnotesize \histnote~Measured on 2026-11-07. The remaining rows are source mechanisms or contract-derived results, not new physical demonstrations; the October demonstrations ran with \code{voice=False}. No economic or environmental benefit is claimed.
\end{flushleft}
\end{table}

Only the first row is available from the box on its own, and it is precisely the throughput result that cannot establish the proposition below. The other five functions exist because sensing, deterministic execution, and orchestration were designed around the retained accelerator. Recycle destroys the silicon; refurbishment restores the old role; repurposing changes only the workload. Re-productization changes the \emph{product role}: an AI box becomes the compute core of a delivered multilingual service robot. The gain is mediated by the three axes below, so it is not the sum of component benchmarks.

\subsection{Three axes of reorganization}
The proposed model has three technical axes:
\begin{itemize}
\item \textbf{Evolution speed.} Assign rapidly changing AI models and orchestration to the old accelerator, while stable execution primitives remain behind a narrow interface. Native sidecars and pinned runtimes contain software-ceiling friction.
\item \textbf{Determinism and safety.} Remove motor-pulse timing from Linux user space. The Xavier sends a semantic command and a finite pulse count; the STM32F407 generates the pulse sequence and stops at completion.
\item \textbf{Peripheral domain.} Reassemble sensing across product domains: a USB microphone array with a recovered protocol, an OpenNI camera used by the person-following runtime, and an automotive DS90UB953/954 video path integrated through V4L2~\cite{ti_ds90ub954}.
\end{itemize}

These axes yield a falsifiable proposition: \emph{the product value of residual silicon depends on the division and orchestration of computation, not only on peak inference throughput}. If identical AI load severely disrupts direct Linux motor timing while the same workload remains useful behind a deterministic execution boundary, then architecture is a causal part of the product result. Throughput benchmarks alone cannot establish this proposition.

\subsection{Evaluation questions}
We evaluate three questions:
\begin{enumerate}
\item[\textbf{Q1}] Can the retired module carry the required AI workloads within its software and memory ceiling?
\item[\textbf{Q2}] Does moving finite motor execution away from Linux address a measurable timing-interference mechanism?
\item[\textbf{Q3}] Does the integrated artifact complete a product-level interaction loop, including its degraded modes?
\end{enumerate}

Q1 requires standalone and concurrent inference measurements. Q2 requires a same-machine A/B comparison, not an analogy to real-time systems. Q3 requires end-to-end traces and a count of failures, not only a photograph or a successful video segment.

\section{A Source-Grounded GEdgeBot Architecture}
\begin{figure*}[t]
\centering
\begin{tikzpicture}[
  font=\small,
  box/.style={draw, rounded corners=1pt, align=center, minimum height=10mm, inner sep=4pt},
  domain/.style={draw, dashed, inner sep=5pt},
  flow/.style={-{Latex[length=2mm]}, thick},
  signal/.style={-{Latex[length=2mm]}, dashed},
  node distance=5mm and 7mm
]
\node[box, fill=blue!7, minimum width=29mm] (mic) {USB microphone array\\DOA frames (0x82)};
\node[box, fill=blue!7, minimum width=29mm, right=of mic] (astra) {Orbbec Astra\\OpenNI color stream};
\node[box, fill=blue!7, minimum width=29mm, right=of astra] (fpd) {AR0143 over\\DS90UB953/954};
\node[box, fill=green!8, minimum width=47mm, below=12mm of astra] (xavier)
  {\textbf{Retired Jetson AGX Xavier}\\Python orchestration; YOLO; ASR; NLLB\\prewarm, wake-up, arbitration};
\node[box, fill=orange!10, minimum width=43mm, right=18mm of xavier] (mcu)
  {\textbf{STM32F407, bare metal}\\frame parsing; finite pulse generation\\automatic stop after each command};
\node[box, fill=orange!10, minimum width=35mm, below=11mm of mcu] (motors)
  {Pan motor and\\three-wheel chassis};
\node[box, fill=purple!8, minimum width=40mm, left=15mm of mcu] (ui)
  {Voice/UI sidecars\\local or configured remote\\LLM/TTS connectors};

\draw[flow] (mic) -- node[above, sloped, font=\footnotesize]{USB CDC} (xavier.north -| mic);
\draw[flow] (astra) -- (xavier);
\draw[signal] (fpd) -- node[above, sloped, font=\footnotesize]{V4L2 integration path} (xavier.north -| fpd);
\draw[flow] (xavier) -- node[above, align=center, font=\footnotesize]{7-byte semantic frames\\UART 9600; no CRC in captured version} (mcu);
\draw[flow] (mcu) -- (motors);
\draw[signal] (xavier) -- (ui);
\node[draw, dotted, fit=(mic)(astra)(fpd), inner sep=4mm, label=above:{Perception interfaces}] {};
\node[draw, dotted, fit=(xavier)(ui), inner sep=4mm, label=below:{Non-deterministic AI and product orchestration}] {};
\node[draw, dotted, fit=(mcu)(motors), inner sep=4mm, label=below:{Deterministic execution domain}] {};
\end{tikzpicture}
\caption{Source-grounded GEdgeBot architecture. The Astra/OpenNI path is used by the captured person-following runtime; the FPD-Link path is a separately documented V4L2 integration and is not assumed to be the camera used in every demonstration. Online connectors are shown explicitly because the current voice assistant is not invariably offline.}
\label{fig:architecture}
\end{figure*}
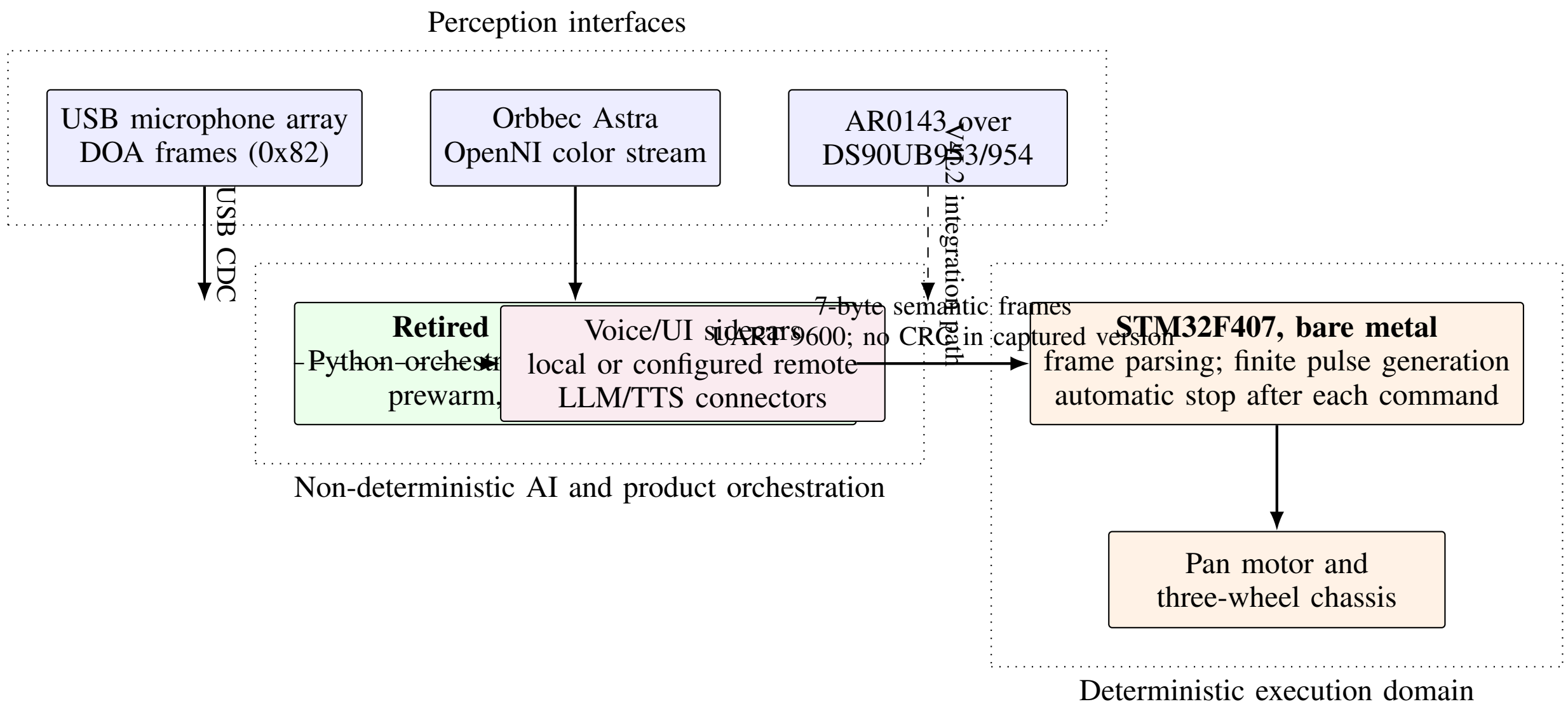

\subsection{Artifact and evidence versions}
Three evidence layers must remain distinct. The October Extended Abstract describes measurements taken on 2026-10-15 through 2026-10-17. The source snapshot audited here was captured on 2026-11-07 from the robot working tree at Git HEAD \code{c25b444b830214f062b08a96dc3b4d7ffa731a0f}, with substantial uncommitted changes; its audited runtime file has SHA-256 \code{d80196b650217cf0e76578855538ab496c}\allowbreak\code{7274be4c74766b5b24dc18dc89b9f5}. The patent specification is a design document, not an execution trace. Consequently, current source proves the presence of mechanisms in the audited version, but does not prove that the identical source produced every October measurement.

The core hardware record is a 32~GB Jetson AGX Xavier on a third-party carrier, an STM32F407 controller, a USB microphone array, an Orbbec Astra camera, and a DS90UB953/954--AR0143 FPD-Link path. The historical software record reports JetPack~5.1.x, L4T~r35, CUDA~11.4, TensorRT~8.5, and Python~3.8. The initial November inspection finds L4T~r35.6.2 with no importable CUDA/cuBLAS/TensorRT stack in the system Python. Inference was subsequently restored without writing to system library directories, using a locally built torchvision, an isolated \code{libcublas}, a \code{LD\_PRELOAD} workaround for scikit-learn's static-TLS failure, a local ChatTTS compatibility patch, and offline model snapshots. This recovery is itself evidence of the software ceiling: a source hash cannot reproduce a runtime, so the experimental record must carry both a clean source snapshot and an executable environment manifest.

\subsection{The overall control plane}
\label{sec:controlplane}
The delivered artifact is not an inference endpoint with a motor cable. Its control plane is a 1,448-line Python orchestrator, \code{RobotControl/robot\_runtime.py} (the audited SHA-256 \code{d80196b650217cf0e76578855538ab496c}\allowbreak\code{7274be4c74766b5b24dc18dc89b9f5}), whose \code{main()} dispatches six CLI modes: \code{mic}, \code{motor-test}, \code{sound-once}, \code{follow}, \code{stdin-wakeup}, and \code{auto-wakeup}. A read-only AST audit of this file counts 13 classes, 69 class methods, 6 module-level functions, 51 CLI arguments, 103 cross-component calls to 62 distinct subsystem methods, 4 daemon-thread creation sites, 2 subprocess launch sites, one threaded HTTP monitor, 7 HTTP client sites, two SIGUSR gates, two 60~s readiness gates, six safety-interlock methods, and 97 control branch sites in 20 categories. All 20 derived checks pass. These are source facts, not runtime measurements.

\begin{table}[t]
\centering
\caption{Overall control plane of the audited runtime (2026-11-07). AST-derived inventory of the captured source snapshot; branch presence is not branch execution, and no physical measurement is claimed.}
\label{tab:control}
\footnotesize
\begin{tabularx}{\columnwidth}{@{}p{27mm}p{24mm}Y@{}}
\toprule
Control element & Derived count & What the combination adds\\
\midrule
CLI modes / dispatch & 6 / 6 & six operating modes around one retained accelerator\\
Concurrent contexts & 4 threads, 2 subprocesses, 1 HTTP server, 7 HTTP clients & voice process, camera capture, monitor, UI/API sidecars\\
Readiness barrier & 2 gates $\times$ 60~s & prewarm hides multi-second model cold starts before wake-up\\
Degraded branches & 97 branch sites, 20 categories & readiness failure, API-vision warning, missing angle, dead zone, dropout, dry-run\\
Safety interlocks & 6 methods & dead zone, two-frame confirmation, rate limits, settle, dry-run default\\
Control boundary & 7 bytes, 3 opcodes & one narrow semantic interface to the STM32F407\\
\bottomrule
\end{tabularx}
\vspace{2pt}
\begin{flushleft}
\footnotesize Counts are re-derivable from the source hash on the laptop and on the Xavier with an identical result fingerprint.
\end{flushleft}
\end{table}

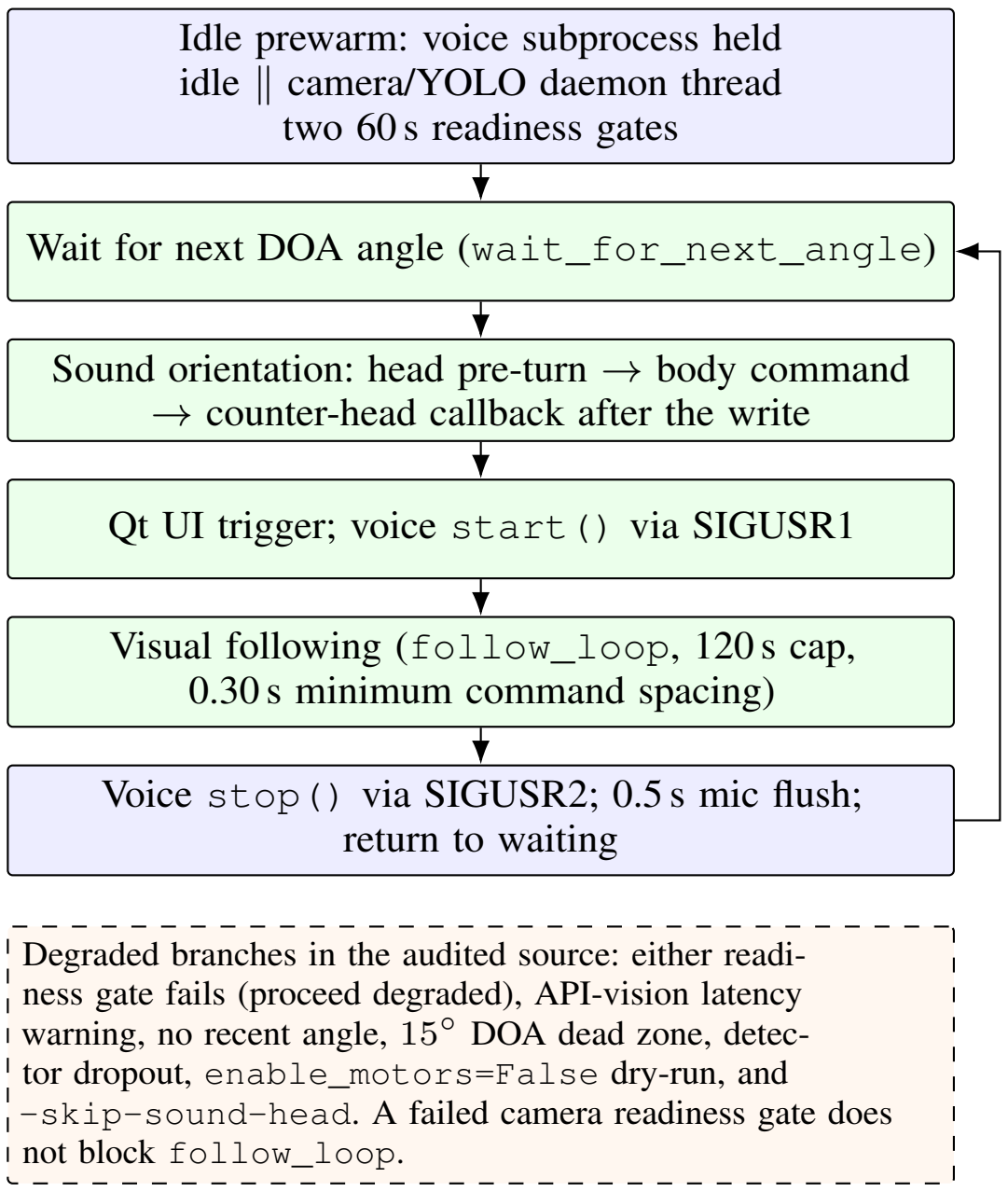
\begin{figure}[t]
\centering
\begin{tikzpicture}[font=\footnotesize, >=Latex,
  box/.style={draw, rounded corners=1pt, align=center, text width=6.5cm,
              inner sep=3pt, minimum height=7mm},
  gate/.style={box, fill=blue!7},
  act/.style={box, fill=green!8},
  deg/.style={draw, dashed, rounded corners=1pt, align=left, text width=6.5cm,
              inner sep=3pt, font=\scriptsize, fill=orange!6},
  node distance=2.6mm]
\node[gate] (pre) {Idle prewarm: voice subprocess held idle $\parallel$ camera/YOLO daemon thread\\ two 60\,s readiness gates};
\node[act, below=of pre] (wait) {Wait for next DOA angle (\code{wait\_for\_next\_angle})};
\node[act, below=of wait] (orient) {Sound orientation: head pre-turn $\to$ body command\\ $\to$ counter-head callback after the write};
\node[act, below=of orient] (voice) {Qt UI trigger; voice \code{start()} via SIGUSR1};
\node[act, below=of voice] (follow) {Visual following (\code{follow\_loop}, 120\,s cap,\\ 0.30\,s minimum command spacing)};
\node[gate, below=of follow] (stop) {Voice \code{stop()} via SIGUSR2; 0.5\,s mic flush;\\ return to waiting};
\draw[->] (pre) -- (wait);
\draw[->] (wait) -- (orient);
\draw[->] (orient) -- (voice);
\draw[->] (voice) -- (follow);
\draw[->] (follow) -- (stop);
\draw[->] (stop.east) -- ++(0.32,0) |- (wait.east);
\node[deg, below=3.5mm of stop] (degraded) {Degraded branches in the audited source: either readiness gate fails (proceed degraded), API-vision latency warning, no recent angle, $15^\circ$ DOA dead zone, detector dropout, \code{enable\_motors=False} dry-run, and \code{--skip-sound-head}. A failed camera readiness gate does not block \code{follow\_loop}.};
\end{tikzpicture}
\caption{Overall control state machine of the audited \code{RobotControl/robot\_runtime.py} \code{auto-wakeup} mode. The prewarm and readiness barrier sit outside the trigger-to-follow path; the box-plus-robot integration is what introduces the concurrent voice process, camera thread, HTTP monitor, microphone array and seven-byte actuator boundary that this state machine orchestrates. Source-derived structure, not a runtime trace.}
\label{fig:control}
\end{figure}

Figure~\ref{fig:control} shows the \code{auto-wakeup} product cycle. The voice subprocess is prewarmed in an idle state, camera and YOLO initialization runs in a daemon thread, and two 60~s readiness gates are evaluated before the wake-word loop. A trigger then performs sound orientation, opens the Qt voice dialog, gates the warm voice process with SIGUSR1, runs a following window capped at 120~s, gates it back with SIGUSR2, and flushes stale microphone angles. Prewarming therefore moves model cold start off the trigger-to-follow critical path; this is the orchestration work that a box-only benchmark cannot exhibit. Three structural findings follow from the same audit. First, \code{follow\_loop(interrupt\_on\_sound=True)} is unreachable from all three audited call sites, so sound re-triggering during following is a reserved but unused capability. Second, \code{MotorController.wait\_hook} is never assigned a callable, making the 0.30~s command spacing and $0.043N+1.2$~s settle waits non-interruptible. Third, the readiness gate checks \code{latest\_status['state']}, whereas \code{follow\_loop} checks only whether the camera-open thread is alive; a failed readiness gate therefore does not block following. These are source-contract findings, not observed failures.

\subsection{A narrow, finite control boundary}
The captured runtime and MCU source implement a seven-byte UART frame at 9,600~bit/s:
\begin{center}
\small
\begin{tabular}{ccccccc}
\toprule
0 & 1 & 2 & 3 & 4 & 5 & 6\\
\midrule
\code{AA} & \code{55} & command & pulse low & pulse high & \code{0D} & \code{0A}\\
\bottomrule
\end{tabular}
\end{center}

For head tracking, bit~15 of the 16-bit pulse parameter encodes direction and the low 15~bits encode magnitude. The MCU receives bytes by interrupt, checks the header and tail, executes the selected finite pulse routine, and calls \code{StopAllMotors()} on completion. The Xavier does not generate individual motor-pulse widths. This is the concrete determinism boundary.

The current frame has no sequence number or CRC. The Extended Abstract's appendix phrase ``header/seq/CRC/tail'' therefore does not describe the audited version. Header/tail checking rejects some malformed frames but does not provide cryptographic integrity or a Hamming-distance guarantee. This limitation matters for any future closed-loop or persistent-velocity controller.

\subsection{Sound-source mapping and counter-turning coordination}
The microphone reader searches for an \code{FF A5} header, checks an eight-bit additive checksum, and decodes little-endian direction-of-arrival (DOA) angles from command~\code{0x82}, subcommand~\code{0x02}. The angle mapper computes
\begin{equation}
\theta = ((\theta_{\mathrm{raw}}+\theta_{\mathrm{offset}}+180)\bmod 360)-180,
\end{equation}
optionally negating $\theta$ for microphone orientation. Angles inside a $15^\circ$ dead zone produce no motion. Otherwise $e=\min(|\theta|,90^\circ)$ and the current code computes
\begin{align}
h &= \operatorname{clip}(\operatorname{round}(e/10),3,H_{\max}),\\
b &= \operatorname{clip}(\operatorname{round}(3e/10),B_{\min},B_{\max}),
\end{align}
where $H_{\max}=15$, $B_{\min}=15$, and $B_{\max}=25$ in the captured defaults.

The orchestration first commands a head pre-turn of $h$ pulses. After a configurable delay---2.0~s in the current defaults---it writes the body command. A callback immediately after that write issues an opposite head command of $h$ pulses. The intended kinematics are that the body and head contributions cancel in world coordinates while the body rotates, leaving the camera directed toward the speaker. This is the host-side intent, not a measured angular trajectory. Moreover, the captured MCU firmware already emits a fixed 15-pulse counter-head sequence inside the blocking \code{TurnBody()} routine. Its UART interrupt only appends received bytes; after the routine returns, \code{ProcessCommand()} resets \code{rx\_idx} without parsing the bytes accumulated during motion. The host's second command therefore has no reliable queued execution path, and the in-routine compensation is fixed at 15 rather than the host-computed $h$. Equal-magnitude cancellation cannot be claimed as an operational end-to-end result. It requires a firmware/host protocol correction followed by synchronized head, chassis, and camera-angle measurements.

\begin{figure}[t]
\centering
\begin{tikzpicture}[font=\footnotesize, x=0.65cm, y=0.65cm]
\draw[-{Latex[length=2mm]}] (0,0) -- (9.5,0) node[right] {time};
\node[anchor=east] at (-0.15,3.3) {Head pan};
\node[anchor=east] at (-0.15,1.9) {Chassis};
\node[anchor=east] at (-0.15,0.6) {Camera ray};
\fill[blue!20] (0.6,3.0) rectangle (3.1,3.6);
\fill[red!18] (3.3,3.0) rectangle (6.5,3.6);
\fill[orange!25] (3.3,1.6) rectangle (6.5,2.2);
\draw[green!50!black, very thick] (0.6,0.3) -- (3.1,0.9) -- (6.5,0.9) -- (7.5,0.9);
\node at (1.85,3.3) {pre-turn $+h$};
\node at (4.9,3.3) {counter-turn $-h$};
\node at (4.9,1.9) {body turn $+b$};
\node[green!50!black] at (4.9,0.55) {approximately fixed on speaker};
\draw[dashed] (3.2,-0.15) -- (3.2,3.9);
\draw[dashed] (6.6,-0.15) -- (6.6,3.9);
\node at (3.2,4.15) {body frame written};
\node at (6.6,4.15) {settle};
\end{tikzpicture}
\caption{Host-side intent of the sound-orientation callback in \texttt{orient\_to\_sound}. The drawing expresses intended kinematic cancellation, not a measured trajectory. The audited MCU's blocking receive/reset logic does not reliably queue the second host command; its in-routine counter-turn is fixed at 15 pulses.}
\label{fig:orchestration}
\end{figure}
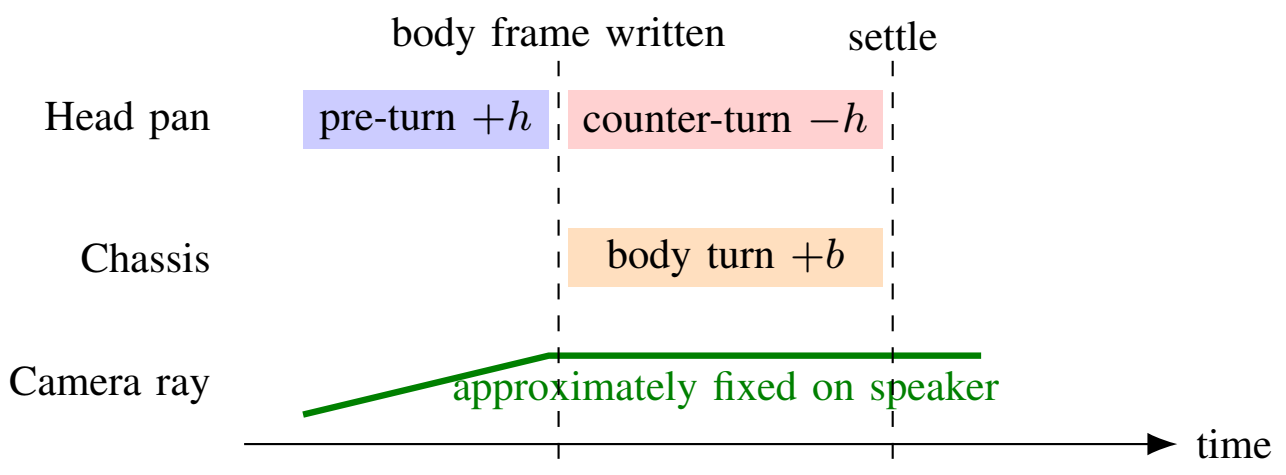

\subsection{Two-level visual following}
The current person-following path opens the Astra color stream through OpenNI and performs YOLO person detection. A follower applies a center dead zone, requires two consistent direction observations, limits head-command rate, and accumulates head pulses by direction. Once the accumulated magnitude reaches 15 and the body interval permits, it adds a finite body-recenter command. The design gives the head a fast, small-motion role and the chassis a slower correction role.

The patent text and current source differ on several constants, including the center zone and small-offset pulse count. The paper therefore uses the audited runtime values when describing implementation and treats the patent as a broader design statement. The FPD-Link driver is a separate integration path; it is not substituted for the Astra camera in the captured \code{CameraFaceTracker} implementation.

\subsection{Voice and translation: bounded offline claims}
The historical measurements cover on-device Whisper-small automatic speech recognition, NLLB-200 translation, and ChatTTS synthesis~\cite{radford2022whisper,nllb2022,chattts2024}. These components can support an on-device translation pipeline. The current conversational assistant, however, also contains cloud-enabled LLM and online TTS connectors alongside local connectors. Thus ``offline ASR/MT/TTS components were measured'' is supportable, while ``every conversational configuration is fully offline'' is not. Each future end-to-end voice benchmark must report the actual ASR, LLM, and TTS backends.

Prewarming is implemented by starting the voice assistant before wake-up, waiting for a readiness marker, and later signaling its transition into listening. Camera/YOLO initialization occurs in parallel. The October demonstration record nevertheless reports \code{voice=False}: the assistant did not become ready within 60~s. We therefore exclude conversational latency from the reported trigger-to-follow timing rather than treating a degraded run as a complete dialogue success.

\subsection{Peripheral registration work}
The FPD-Link implementation creates serializer-side I$^2$C clients from device-tree children and connects the sensor/deserializer/NVCSI graph. This is more than a cabling choice: client creation, device-tree node association, and endpoint references determine whether the sensor driver can probe and register a V4L2 subdevice. The current source supports a mechanism-level account of this work. It does not, by itself, establish video throughput, long-term flex reliability, or use of this camera in the October following demonstrations.

\section{Experimental Method and Evidence Provenance}
\subsection{Historical October measurements}
The historical record reports the following procedures. YOLOv8n was measured at 320 and 640 input sizes with FP32 CUDA execution, with $n=60$ per run~\cite{jocher2023yolov8}. Whisper-small was measured using audio real-time factor (RTF), NLLB-200 with greedy FP16 decoding, and ChatTTS with synthesis RTF. A three-model configuration loaded YOLO, Whisper, and NLLB together to assess memory residency and contention.

The control experiment compared a 50~ms user-space frame period with and without concurrent YOLO@640 load, using 200 frames per condition. The historical summary reports interval standard deviation, p99 timing deviation, and maximum slip under idle 15~W, AI-loaded 15~W, and AI-loaded MAXN conditions. Fault-injection cases stopped frame transmission, closed the serial port, and injected malformed protocol input.

The end-to-end demonstration comprised six runs in five capture windows. The measured chain was microphone trigger, DOA-guided turn, assistant activation, and visual following. Because the voice assistant was not ready, success means completion of the trigger-to-follow chain, not a verified spoken dialogue.

All October values in this draft are \emph{transcribed historical values}. Their scripts and raw JSON logs were not found in the inspected backup, old session stores, or current repository. They must be checked against original artifacts or rerun before a submission claims independent reproduction.

\subsection{New November validation}
We captured the current runtime without modifying the robot repository and wrote tests only for pure functions and parsers. Seventeen tests cover motor framing, direction-bit encoding, pulse clipping, microphone checksums and DOA decoding, sound-angle mapping, visual-follow confirmation, rate limiting, and accumulated body correction. The same suite passed on macOS and on the Xavier's Python~3.8 environment. No device was opened by these tests.

For the new timing run, the Xavier sent 201 production-format frames at a 50~ms absolute schedule to a POSIX pseudo-terminal, yielding 200 intervals. The receiver returned all 201 frames with an exact byte-sequence match. This run used \code{os.write} but not \code{tcdrain}, a real UART, or a motor. It measures scheduling and user-space write behavior only; a later November run uses the real motor port with inert frames. The machine was already in MAXN; the script did not change power mode.

A separate six-check static audit compares the host callback with the captured MCU routine. It checks the fixed 15-pulse in-routine counter-turn, blocking delays, interrupt-only byte buffering, and receive-index reset after motion. These checks establish a structural queue/execution gap; they are not presented as a dynamic observation of a dropped physical command.

For intervals $\{t_i\}$ in milliseconds, we report population standard deviation and the nearest-rank p99 of $|t_i-50|$. This definition is included because the absent original script prevents confirmation that every historical statistic used an identical estimator.

\subsection{New November device measurement campaign}
A second November campaign was run on the same Xavier after the isolated stack described in Section~V was assembled. It comprises three reconstructed AI runs at MAXN and one at 15~W; source-aligned runs with three repeats per power mode; five paired idle/AI-loaded control-frame runs per power mode with a CPU-only control; Whisper and NLLB parameter sweeps; a fresh-process cold-start control; ChatTTS synthesis; and a nine-case offline protocol guard. Every run pins the model snapshots, the audited source commit \code{c25b444}, 60 YOLO iterations, five NLLB runs per length, and the power mode, and records per-frame timestamps. These are \emph{new measurements}, not recovered October logs; they are reported in a parallel table and never pooled with the historical column. A further real-UART run writes inert frames to the motor port without actuating it (Section~V). F1/F2 injection during motion, the six-run demonstration, and logic-analyser capture of the MCU pulse pins were not performed; these cases remain behind an explicit on-site confirmation guard.

\subsection{Real-UART no-motion command-phase measurement}
The motor link is present and usable: \code{/dev/ttyUSB0} is a QinHeng HL-340 bridge at 9,600~bit/s, no process holds the port, and a read-only probe of both serial devices returned no bytes. The firmware emits only a startup banner, with neither a per-command acknowledgement nor per-pulse timestamps. To measure the command start phase on the real transport without moving the robot, three runs of 201 inert seven-byte frames were written at a 50~ms absolute schedule in each of the idle and YOLO@640-loaded conditions at MAXN. The frame uses \code{CMD}=\code{0x00}, for which the audited \code{ProcessCommand()} switch has no case (it dispatches \code{0x01}--\code{0x09} only). The firmware snapshot is parsed before the port is opened, and the write requires an explicit confirmation flag. We record \code{os.write()} and \code{tcdrain()} completion times and apply the same interval estimator as above. No MCU acknowledgement is observed, so reception is not claimed. A seven-byte frame needs about 7.29~ms on the wire at 9,600~bit/s, whereas \code{tcdrain()} returns in about 0.15~ms; USB bulk transfer therefore hides the true arrival instant from the host. This run measures command emission only, not the pulse train on the MCU pins.

\subsection{Trace-driven boundary analysis}
To test Q2 without a physical actuator, 6,000 measured 50~ms frame intervals from 30 control-frame runs drive two execution organisations: a counterfactual in which Linux generates the pulses directly, and the shipped seven-byte finite-pulse offload. Execution-side quantities are derived from the audited firmware contract---the \code{TurnBody()} loop, \code{SPEED\_NORMAL}=35~ms, an 8~ms inter-pulse gap, and \code{StopAllMotors()}---rather than observed with a logic analyser. All 30 runs report an exact received frame sequence and no physical motor writes, which is the precondition for the semantic-command argument. This is an analytical result and is labelled as such wherever its numbers appear.

\subsection{Software-in-the-loop product-loop replay}
A third November component makes Q3 auditable without imputing a physical demonstration. It imports the audited \code{robot\_runtime.py} snapshot (commit \code{c25b444}) and executes its real decision and frame-building functions: \code{AngleMapper.plan\_sound\_orient}, \code{RobotRuntime.orient\_to\_sound}, \code{VisualFollower.update}, and \code{MotorController.pan\_head}/\code{turn\_body}. Microphone, camera, and serial sources are replaced by in-memory event streams, and every device-opening method in the loaded module is replaced by a guard that raises, so no frame can reach hardware. A virtual clock advances only through \code{sleep} calls already present in the source.

The six transcribed DOA turn angles seed 30 traces; face-position streams are synthetic probes paced at the transcribed 17.7~FPS, not recorded detector output. Five conditions cover nominal operation, \code{voice=False}, detector dropout, dead-zone DOA, and a deliberately injected camera-not-ready failure. Completion means the trigger-to-follow chain only. The replay claims no wall-clock measurement, no physical actuation, and no conversational success. The same script ran on the laptop and on the Xavier with identical source hashes and an identical result fingerprint.

\subsection{Full-runtime control-state replay}
A fourth November component makes the \emph{overall} control plane auditable rather than only one decision chain. It loads the same audited module and executes its real \code{main()} CLI dispatcher and all six \code{run\_*} methods, including the complete \code{run\_auto\_wakeup} state machine. Peripherals are replaced by recording stubs: microphone events, camera face streams, serial sinks, an HTTP monitor, and a SIGUSR-gated voice process. Every physical-opening method in the loaded module is replaced by a guard that raises and increments a counter, HTTP calls resolve to an in-memory 200 response, and the module's \code{time} is a virtual clock. The real \code{AngleMapper}, \code{VisualFollower}, \code{MotorController} and seven-byte frame builders still execute. Fourteen scenarios cover all six modes, ten complete auto-wakeup cycles, eight degraded branches and a deliberate dry-run case; their event logs, frame contracts and a ten-check cross-audit are retained in JSON. The same script ran on the laptop and on the Xavier with identical source hashes and an identical result fingerprint. It claims no physical actuation, no wall-clock latency and no dialogue success.

\section{Results}
\subsection{Q1: Historical AI feasibility and contention}
Table~\ref{tab:ai-results} reproduces the October summary. The central architectural result is not peak FPS; it is that the three required models can reportedly coexist in 1,782~MB of GPU memory, with YOLO latency degrading from 37.6 to 59.0~ms under contention while remaining functional for the product loop. This supports a feasibility claim for one configured artifact, not a general performance ranking.

\begin{table}[t]
\centering
\caption{Historical single-prototype AI results (October 2026). Values are transcribed, not independently re-derived in this draft.}
\label{tab:ai-results}
\small
\begin{tabularx}{\columnwidth}{@{}p{35mm}Y@{}}
\toprule
Workload & Reported result\\
\midrule
YOLOv8n@320, 15~W & 26.6 FPS; mean 37.6~ms; p99 84.0~ms\\
YOLOv8n@640, 15~W & 16.3 FPS; mean 61.2~ms\\
Whisper-small, 10~s audio & RTF 0.091; 2~s clip RTF 0.44\\
NLLB-200, short/mid/long & 0.42/1.02/1.55~s in the single-model table\\
ChatTTS, long text & RTF 2.51; not real time\\
Three models resident & 1,782~MB GPU memory\\
YOLO under contention & mean 59.0~ms; p99 173.5~ms\\
\bottomrule
\end{tabularx}
\end{table}

\begin{table}[t]
\centering
\caption{New November 2026 measurements on the same Xavier, using audited source parameters ($n=3$ repeats per power mode). An independent batch: not a reproduction of the October values and never pooled with Table~\ref{tab:ai-results}.}
\label{tab:nov-results}
\footnotesize
\begin{tabularx}{\columnwidth}{@{}p{32mm}YY@{}}
\toprule
Metric & MAXN & 15~W\\
\midrule
YOLOv8n@320 (ms) & $23.40\pm6.23$\histnote & $18.65\pm0.03$\\
YOLOv8n@640 (ms) & $22.14\pm1.24$ & $24.44\pm0.26$\\
Whisper 10~s RTF & $0.159\pm0.003$ & $0.215\pm0.008$\\
NLLB short, beam~5 (s) & $0.735\pm0.010$ & $0.762\pm0.019$\\
NLLB short, greedy (s) & $0.463\pm0.004$ & $0.480\pm0.015$\\
ChatTTS long-text RTF & 2.982 & ---\\
F3 protocol guard & \multicolumn{2}{c}{9/9 offline cases}\\
\bottomrule
\end{tabularx}
\vspace{2pt}
\begin{flushleft}
\footnotesize \histnote~Cross-repeat CV 27\% (per-run means 21.99, 16.57, 31.64~ms); the per-run values, not the pooled mean, should be reported for this workload.
\end{flushleft}
\end{table}

Table~\ref{tab:nov-results} is a new, independently auditable batch. Two of its findings change how the historical AI numbers must be read. First, the initial reconstructed Whisper run reported an RTF near 2.27 because a fresh FP32 pipeline pays a one-off CUDA/cuDNN initialisation of 20.9~s at MAXN and 23.2~s at 15~W; a fresh-process cold-start control reproduces RTF 2.086 and 2.317, while steady state falls to 0.138 and 0.212 and the FP16 prewarmed 10~s RTF reaches 0.100 and 0.151. The historical 0.091 is therefore a setup- and estimator-dependent quantity, not evidence of a hardware regression. Second, the audited service uses beam-5 NLLB decoding while the historical summary describes greedy; both were measured on the same model, and the greedy control is closer to the transcribed 0.42/1.02/1.55~s. Neither result identifies the lost October prompt or audio, so the 0.42/1.33~s conflict remains unresolved.

The power-sensitivity table reports YOLO@320 increasing from 26.6 to 28.0~FPS between 15~W and MAXN, and Whisper 10~s RTF improving from 0.091 to 0.032. It also reports NLLB short-text latency as 1.33~s at 15~W and 0.66~s at MAXN. The same abstract's single-model table gives 0.42~s for short text at 15~W. Without prompts, run identifiers, and raw logs, this discrepancy cannot be resolved. We therefore do not use the reported 2.0$\times$ NLLB power scaling as evidence for a quantitative conclusion.

\subsection{Q2: Timing interference and the execution boundary}
The historical A/B summary reports p99 timing deviation rising from 0.24~ms in the idle 15~W condition to 49.93~ms under concurrent AI load, a ratio of approximately $208\times$. Under MAXN with AI load, the reported value is 49.92~ms. In that record, higher clocks alone do not remove the interference. This is consistent with the architectural hypothesis, but the missing raw data prevents testing distributional assumptions or reproducing the ratio.

\begin{table}[t]
\centering
\caption{Control-timing records. The November run is a different, no-motion protocol and must not be pooled with October data.}
\label{tab:timing}
\small
\begin{tabularx}{\columnwidth}{@{}p{27mm}rrr@{}}
\toprule
Condition & Std (ms) & p99 dev. (ms) & Max (ms)\\
\midrule
Oct., idle, 15~W\histnote & 0.063 & 0.24 & ---\\
Oct., AI, 15~W\histnote & 13.37 & 49.93 & 212\histnote\\
Oct., AI, MAXN\histnote & 11.41 & 49.92 & 182\histnote\\
Nov., idle, MAXN, pty & 0.216 & 0.246 & 52.136\\
\bottomrule
\end{tabularx}
\vspace{2pt}
\begin{flushleft}
\footnotesize \histnote~Historical transcription. The October maximum entries are described as single-frame slip and may not use the same definition as the November maximum interval.
\end{flushleft}
\end{table}

The new November run provides a smaller but independently auditable result. Over 200 intervals, the mean interval is 49.999996~ms, population standard deviation is 0.216396~ms, and p99 absolute deviation is 0.246103~ms. The largest interval is 52.136408~ms; the following interval is 47.846319~ms because the absolute scheduler catches up after a delayed frame. The mean \code{os.write} time is 0.029837~ms and all 201 frames are received exactly.

The close p99 values in the two idle records are interesting, but not an equivalence result: power mode, transport, drain behavior, and implementation differ. An AI-loaded comparison was therefore run on the restored isolated stack, five paired runs per power mode plus a CPU-only control. It does \emph{not} yield a repeatable factor: MAXN paired p99 ratios span 0.14--$38.5\times$ and 15~W spans 0.88--$8.55\times$, idle runs themselves contain scheduler spikes, and the CPU-only control does not reproduce the largest slips. We therefore report no new ratio and do not claim a reproduction of $208\times$. The 15~W baseline is additionally contaminated by resident camera and streaming services and by a four-core \code{schedutil} configuration; no service was stopped for this study.

Because those runs used a pseudo-terminal, the paired comparison was repeated on the real motor UART with inert frames (Table~\ref{tab:uart}). The link is online, but this does not recover the $208\times$ result: the UART carries a single command per motion, whereas the historical statistic describes the cadence of pulses generated on MCU GPIO pins, which no host-side timestamp can observe. Over three runs per condition, the idle p99 spans 0.093--0.480~ms and the AI-loaded p99 spans 0.095--0.104~ms; AI load again produces no repeatable degradation, and the largest deviation occurs in an idle run. Table~\ref{tab:uart} is a separate November measurement and is not pooled with Table~\ref{tab:timing} or with the October column.

\begin{table}[t]
\centering
\caption{Command start-phase deviation on the real motor UART and on the pty transport, all November 2026 no-motion measurements with the same estimator. Real-port frames are inert; the historical pulse-train statistic is not reproduced here.}
\label{tab:uart}
\footnotesize
\begin{tabularx}{\columnwidth}{@{}Ylcr@{}}
\toprule
Transport & Load & $n$ & p99 range (ms)\\
\midrule
Real \code{/dev/ttyUSB0} & Idle & 3 & 0.093--0.480\\
Real \code{/dev/ttyUSB0} & YOLO@640 & 3 & 0.095--0.104\\
pty (Section~V) & Idle & 5 & 0.051--0.871\\
pty (Section~V) & YOLO@640 & 5 & 0.064--2.908\\
\bottomrule
\end{tabularx}
\end{table}

Rather than pursue a fragile single number, we test the architectural proposition directly. Driving two execution organisations with the same 6,000 measured 50~ms frame intervals separates two quantities that the phrase ``remove motor-pulse timing from Linux'' conflates. With direct Linux pulse generation, host deviation transfers 1:1 into the pulse train; the measured host p99 ranges from 0.083~ms at idle MAXN to 5.101~ms under 15~W AI load. With finite-pulse offload, that transfer is zero \emph{by contract}: the MCU emits $N$ pulses on its local clock and then calls \code{StopAllMotors()}. The boundary does not remove the command start-phase deviation, which still equals the measured host value. Architecture thus isolates pulse-train cadence and runaway motion, not command latency. These ratios are contract properties rather than empirical estimates, and MCU-local clock drift was not measured with a logic analyser. The real-UART run narrows but does not close this gap: it replaces a pty with the production transport for the command start phase, while the pulse train itself remains unobserved.

A static source check sharpens the safety bound. \code{ProcessCommand()} passes an unclamped 16-bit pulse field straight to \code{TurnBody()}; the limit of 25 body pulses is enforced only by the host-side clamp. With \code{SPEED\_NORMAL}=35~ms and an 8~ms gap, residual motion after the host stops is at most 1.08~s for a legitimate command but 46.97~min for the largest frame value. Since the frame carries neither a sequence number nor a CRC, a corrupted pulse field is undetected, while a persistent-velocity command would be unbounded because no command timeout or watchdog exists. Finite-pulse execution removes one runaway state, but its bound is a host-side policy, not an MCU guarantee.

Historical fault-injection cases report that stopping frames or closing the serial port leaves no persistent motion, and malformed frames produce no spurious motion. This behavior is consistent with finite-pulse execution. These cases require the wheels to be moving and a person present at the machine: inert no-motion frames cannot falsify persistent motion, so the injection cases are deliberately left unclaimed here even though the serial link is available. However, no hardware watchdog or command timeout is claimed. A future persistent-velocity design would need a different safety analysis.

\subsection{Q3: Product-loop existence evidence}
The October record reports six of six completed trigger-to-follow runs, with mean trigger-to-follow latency 4.35~s, reported $\pm0.19$~s spread, p50 4.42~s, minimum 3.95~s, and maximum 4.43~s. The recorded DOA turn angles are $+90^\circ$, $-53^\circ$, $+104^\circ$, $+110^\circ$, $-97^\circ$, and $+114^\circ$. Three follow segments lasted 120~s, two lasted 30~s, and one ended with its capture window. Live perception is reported at approximately 17.7~FPS and 37~ms inference.

These observations support the existence of a working trigger-to-follow chain on one prototype. They do not establish a 100\% fleet success rate. The sample is small, capture windows are not fully independent deployments, and the voice subsystem was in a documented degraded mode. No conversational success rate or dialogue latency can be inferred.

To make the product-loop claim independently auditable without a physical rerun, the software-in-the-loop replay of Section~V drives the same audited functions with synthetic events. Table~\ref{tab:replay} reports 30 traces: the trigger-to-follow chain completed in 24/24 runnable cases, while a deliberately injected camera-not-ready condition completed 0/6. All 600 emitted seven-byte frames passed the header/tail, opcode, and host-clamp contract, and no physical device was opened.

\begin{table}[t]
\centering
\caption{Software-in-the-loop product-loop replay (2026-11-07): 30 traces driven by synthetic events and a virtual clock. Not a physical demonstration and not a reproduction of the October runs.}
\label{tab:replay}
\footnotesize
\begin{tabularx}{\columnwidth}{@{}p{30mm}YY@{}}
\toprule
Condition & $n$ / completed & Orient budget (s)\\
\midrule
Nominal & 6 / 6 & $4.211\pm0.144$\\
Voice degraded & 6 / 6 & $4.211\pm0.144$\\
Detector dropout & 6 / 6 & $4.211\pm0.144$\\
DOA dead zone & 6 / 6 & 0\\
Camera not ready & 6 / 0 & $4.211\pm0.144$\\
\bottomrule
\end{tabularx}
\vspace{2pt}
\begin{flushleft}
\footnotesize Mean $\pm$ population stdev over the six seed angles. Virtual-clock budgets, not wall-clock timings; the dead-zone condition skips orientation by design.
\end{flushleft}
\end{table}

Voice-degraded traces are bit-identical to nominal ones because the audited source starts the voice subprocess in parallel with, rather than inside, the trigger-to-follow critical path. This is a source-structure result: it explains how the October \code{voice=False} record can coexist with six completed trigger-to-follow runs, but it is not a dialogue success rate. Two further findings are architectural. Four of the six transcribed DOA angles exceed $90^\circ$ and saturate to the same 9-head/25-body pulse plan. The 3.888--4.275~s orientation budget is dominated by fixed host waits---a 2.0~s head--body delay, 0.30~s command spacing, and $0.043N+1.2$~s settling---rather than by inference, so product-level latency predictability comes from orchestration rather than peak throughput. The replay also emits 24 counter-head frames for which the audited MCU firmware has no reliable queue, consistent with Section~VI. These are virtual-clock budgets and source contracts, not wall-clock measurements; they are reported in parallel with, never pooled with, the historical six-run record.

\subsection{Q3 extension: overall control-state coverage}
Table~\ref{tab:fullreplay} reports the full-runtime replay. All 14 scenarios completed across all six CLI modes, including ten complete auto-wakeup cycles and 14 following windows. The 54 emitted seven-byte frames passed the header/tail, opcode and host-clamp contract with zero errors, and the guarded physical-opening methods were invoked zero times; all ten cross-checks pass.

\begin{table}[t]
\centering
\caption{Full-runtime control-state replay (2026-11-07): 14 offline scenarios executing the real six-mode dispatcher. Not a physical demonstration and not a reproduction of the October runs.}
\label{tab:fullreplay}
\footnotesize
\begin{tabularx}{\columnwidth}{@{}p{31mm}p{23mm}Y@{}}
\toprule
Measure & Result & Interpretation\\
\midrule
Scenarios / modes & 14/14, 6/6 & every audited entry point executes\\
Auto-wakeup cycles & 10 & prewarm $\to$ trigger $\to$ orient $\to$ voice $\to$ follow $\to$ flush\\
Frames / errors & 54 / 0 & seven-byte contract and host clamps hold\\
Physical / logical opens & 0 / 33 & guards fired 0 times; stubs recorded calls\\
State transitions & 48 distinct & ordered, machine-readable control trace\\
Degraded branches & 8 exercised & readiness, API warning, dead zone, dropout, dry-run, no angle, skip-head\\
Cross-checks & 10/10 & prewarm/gate ordering and SIGUSR gating included\\
\bottomrule
\end{tabularx}
\vspace{2pt}
\begin{flushleft}
\footnotesize Virtual-clock budgets, not wall-clock timings. The laptop and Xavier runs produced an identical result fingerprint on the identical source hash.
\end{flushleft}
\end{table}

Three findings bear on the re-productization proposition. First, voice prewarm and both 60~s readiness gates precede the first DOA trigger in every auto-wakeup scenario, and SIGUSR1/SIGUSR2 gating is correctly ordered: the integration schedules multi-second model cold starts into idle time, so they do not appear in the trigger-to-follow path. Second, the eight degraded branches are exercised by real source branches rather than test scaffolding; with the default \code{enable\_motors=False} the source writes zero serial frames. Third, \code{interrupt\_on\_sound=True} occurs zero times, matching the static audit, and a false camera readiness gate still enters \code{follow\_loop} because the two code paths use different readiness predicates. Both are source-structure findings, not runtime faults. One-cycle virtual budgets are 3.475--5.475~s and are dominated by the 2.0~s head--body delay, 0.30~s command spacing and $0.043N+1.2$~s settling, again indicating that product-level latency is set by orchestration rather than peak inference throughput. These are source-derived budgets, not wall-clock measurements, and are never pooled with the historical 4.35~s record.

\subsection{New source-contract validation}
All 17 contract tests pass on the Xavier. In addition to framing and parsing, they expose version-specific behavior that a high-level description hides. For example, a $120^\circ$ input saturates to $90^\circ$, producing nine head pulses and 27 calculated body pulses before clipping; the current default body limit reduces the latter to 25. The patent example obtains 27 by using a larger limit. The six-check firmware audit passes as a characterization of the captured source and reveals a more substantive issue: the host emits the planned second counter-head frame, but the MCU's blocking routine and receive-index reset provide no reliable queue for it. This finding does not assert that an oscilloscope observed a dropped frame; it shows that the present source cannot be used as sufficient evidence for the patent's idealized equal-magnitude sequence.

Other discrepancies do not necessarily indicate an implementation error; they indicate that parameters and protocol revisions must be versioned as experimental data. Table~\ref{tab:versions} records the most consequential examples.

\begin{table}[t]
\centering
\caption{Design-document and captured-source differences. Different entries may describe different revisions; none is silently treated as the October execution record.}
\label{tab:versions}
\footnotesize
\begin{tabularx}{\columnwidth}{@{}p{22mm}p{28mm}Y@{}}
\toprule
Item & Design/historical text & Captured source and implication\\
\midrule
Motor integrity & Abstract: sequence and CRC & Seven bytes; no sequence/CRC; header/tail only\\
Turn opcodes & Patent: left \code{05}, right \code{06} & Runtime: left \code{01}, right \code{02}\\
Head--body delay & Patent: 0.4~s & Runtime default: 2.0~s\\
Sound body limit & Patent example: up to 60 pulses & Runtime default maximum: 25 pulses\\
Small follow pulse & Patent: 3 pulses & Runtime selects 5 and enforces a minimum of 5\\
Counter-turn path & Patent: equal planned opposite pulse count & MCU uses fixed 15 in routine; buffered host frame can be discarded\\
Following camera & Multiple sensors in product record & Astra/OpenNI is explicit in the captured follow loop\\
\bottomrule
\end{tabularx}
\end{table}

\section{Discussion}
\subsection{What the evidence says about re-productization}
The strongest synthesis across Q1--Q3 is that residual accelerator value is mediated by system organization. AI feasibility alone would leave open how a general-purpose OS can safely operate the robot. Control isolation alone would leave open whether the old accelerator can run the product workload. The artifact is informative because it joins these requirements and exposes the interfaces between them.

The historical A/B result is consistent with a causal role for architectural separation: the compute module did not change between the interfering and non-interfering control organizations. Yet the present draft cannot reconstruct that comparison from raw data, and the new no-motion result covers only idle behavior. The appropriate conclusion is therefore a supported architectural hypothesis with a concrete replication procedure, not a completed independent validation of every historical number.

Table~\ref{tab:integration} sharpens this synthesis: one of six product functions is available from the module alone, while the other five require sensing, a deterministic execution boundary, and orchestration designed around it. Section~\ref{sec:controlplane} identifies where that orchestration lives: a six-mode control plane with four daemon threads, two subprocess sites, an HTTP monitor, two readiness gates and 97 control branches. Table~\ref{tab:fullreplay} shows it executing end-to-end offline. Value is created at the interface between the old compute core and the new product organisation, which is why a benchmark-only reuse study would miss it. The joint character of the gain also explains why neither the AI feasibility results nor the timing results are sufficient on their own: each covers one side of the same architectural division.

\subsection{Engineering labor is part of the architecture}
The microphone checksum/DOA parser, the six-mode dispatcher, the prewarm/readiness protocol with SIGUSR1/SIGUSR2 gating, finite UART commands, and FPD-Link client/device-tree association are not incidental implementation trivia. They determine whether the compute module can participate in the product at all. Re-productization provides a way to make this labor visible: the unit of reuse is the accelerator, but the unit of design is the system.

This perspective also avoids a misleading ``only the SoC is old'' slogan. Project documents describe both retained mechanical/motor-drive elements and newly designed subsystems. The model does not require a particular reused fraction. It requires that the retained compute module be placed in a deliberately redesigned product organization.

The physical side of that redesign was carried out in collaboration with the University of Science and Technology Beijing (USTB): the collaboration reinforced the delivered platform and refreshed its hardware so that the retained compute module could continue to serve as the compute core of a mobile product. We state this as an engineering-contribution record rather than as an experimental result. No before/after reliability, thermal, vibration, or mechanical data were collected for the reinforcement work, no such quantity appears in Sections~V--VI, and none of the reported measurements or source audits is attributed to it. The collaboration therefore supports the qualitative account of re-productization labor, not a quantified improvement claim.

\subsection{Safety and protocol limitations}
Finite-pulse commands remove persistent-PWM runaway as a failure state, but the analysis in Section~VI shows that the bound is enforced by the host clamp rather than by the firmware, which accepts any 16-bit pulse count. They therefore do not prove bounded stopping distance, obstacle safety, or mechanical limit enforcement. The seven-byte frame lacks CRC and sequencing, so a corrupted pulse field is undetectable. The counter-turning mechanism still lacks a measured angular-error trace. Three concrete improvements follow directly: add integrity and command-age checks, clamp the pulse count inside the firmware, and measure synchronized angles under a documented mechanical safety setup.

\subsection{Environmental and product interpretation}
\label{sec:interpret}
The e-waste context motivates the work but does not quantify its benefit. A defensible lifecycle comparison would require the reused module's allocation rule, new carrier and peripherals, energy use, expected service life, and the counterfactual replacement product. Likewise, the study provides no sales, deployment, or failure-rate evidence. What the study does establish is a \emph{product-level capability gain}: six functions from one retained compute module plus a redesigned system organisation, as inventoried in Table~\ref{tab:integration}. We do not assign that gain a monetary or carbon value. Re-productization is therefore presented as an architectural category whose environmental and economic evaluation remains open.

\section{Limitations and Reproduction Plan}
This study is limited by one prototype; unavailable October raw logs and benchmark scripts; uncommitted differences between Git HEAD and the audited working tree; a current AI software environment that differs from the historical environment; inconsistent NLLB summary values; protocol and parameter differences across documents; degraded voice in the end-to-end demonstrations; the host--MCU counter-turn queue mismatch; a trace-driven boundary analysis whose execution-side quantities are contract-derived rather than physically observed; an MCU pulse bound enforced only by the host; a product-loop replay and a six-mode full-runtime replay, both built from synthetic scenario events and a virtual clock rather than observed sensor traces, which establish source-level behaviour but no physical completion rate; and the absence of measured mechanical trajectories for counter-turning coordination.

The minimum reproduction sequence is:
\begin{enumerate}
\item Recover the October system image, dependency lockfiles, scripts, and JSON logs, or declare them unavailable and rerun every required condition.
\item Preserve an executable environment manifest and a clean source snapshot rather than relying on a dirty Git working tree.
\item Repeat standalone and concurrent AI measurements with fixed inputs, prompts, warmup, iteration counts, and power modes.
\item Capture the MCU pulse pins with a logic analyser under the same AI load and statistic definitions. The motor UART is already online and was exercised with inert frames, so the missing element is waveform capture at the actuator, not link availability.
\item Run fault and motion tests only with the robot mechanically secured and an operator present.
\item Correct and dynamically verify the counter-turn command queue, unify the camera readiness predicate used by the readiness gate and the follow loop, then measure actual head/body/camera angles during sound orientation; report the selected local or remote backend for every voice stage.
\item Repeat on multiple modules and carriers before making cross-device reliability claims.
\end{enumerate}

The November measurement scripts, per-frame JSON traces, isolated environment manifest, static-audit results, trace-driven analyses, and the seven \LaTeX{} source files of this manuscript are released at \url{https://github.com/gravity-link/gedge_bot_paper} (release \texttt{v1.0.0}, archived at \url{https://doi.org/10.5281/zenodo.24152092}), under MIT for code and CC BY 4.0 for data and documentation. The whitelist-only release contains 24 Python scripts, 85 measurement and analysis JSON records, nine on-device runner shell scripts, and the manuscript sources, byte-identical to the artifacts used here; each JSON record embeds the SHA-256 hash of the script that produced it, and the repository ships a checker that re-verifies those hashes. Private patent materials, the proprietary runtime and STM32 firmware snapshots, raw audio captures, and internal reports are excluded, with the complete inclusion/exclusion record published in the repository. The October raw logs and benchmark scripts remain unavailable and are not part of this release, so all October values in this paper remain transcribed historical values with explicit evidence status. Patent documents are treated as private design materials and are not needed as public references for the architectural claims. Publishing these artifacts does not convert replay-only or contract-derived results into physical measurements; the on-device conditions listed above are still required for physical reproduction.

\section{Conclusion}
GEdgeBot demonstrates why the end of a compute module's product-support trajectory need not be treated as the end of its architectural usefulness. Re-productization reframes reuse as a redesign of computation, control, and perception around a retained accelerator. The delivered robot makes the joint value visible: the box alone runs models, while the box together with sensing, a deterministic execution boundary, and orchestration becomes a multilingual service product. The delivered robot and historical measurements provide an existence case; the new source audit, the independently rerun AI campaign, a real-UART no-motion command-phase measurement, the trace-driven isolation analysis, a 30-trace software-in-the-loop product-loop replay, and a 14-scenario six-mode full-runtime control-state replay provide a narrower but reproducible foundation. The central hypothesis---that system organization, not residual compute alone, determines recovered product value---is now testable in a specific form: the finite-pulse boundary isolates pulse-train cadence and bounds motion, while leaving command start-phase jitter untouched. The motor UART is online and was measured with inert frames, so link availability is not the obstacle. Completing the physical test instead requires capturing the MCU pulse pins with a logic analyser under the same AI load, running fault injection while the robot is moving under a documented safety setup, preserving the field data, correcting the counter-turn queue, clamping the pulse count in firmware, and validating the mechanical and conversational paths with the same rigor as the control protocol. The historical $208\times$ ratio remains an unreplicated transcription and is not used as evidence for the hypothesis.

\section*{Acknowledgment}
Author contributions are stated explicitly because this draft separates measured results from engineering work. H. Luo designed the study, reconstructed the evidence base, and drafted the manuscript; S. Tie (University of Science and Technology Beijing) led the mechanical reinforcement and hardware refresh of the delivered platform; J. Wang (Beijing Data Chain Technology Co., Ltd.) led testing and validation of the delivered system, including operating-workflow validation, safety-related testing, and interaction testing. These are engineering-contribution records, not additional experimental evidence: neither the platform work nor the engineering tests supply measurements, source audits, or analyses reported in Sections~V--VI, and they do not constitute the on-site motion-state fault injection, the six-run physical demonstration, or the MCU pulse-pin waveform capture that remain open in the reproduction plan. The authors thank the collaborators and endorsers who reviewed the project concept. AI-assisted drafting and source-analysis tools were used; no AI system generated the physical measurements. The authors remain responsible for verifying the final submission and for all technical claims.

\bibliographystyle{IEEEtran}
\bibliography{references}
\end{document}